\documentclass[10pt,journal,compsoc]{IEEEtran}
\usepackage{cite}
\ifCLASSINFOpdf
   \usepackage[pdftex]{graphicx}
\else
\fi
\usepackage{amsmath}
\usepackage{amssymb}
\usepackage{tabulary,xfrac,enumitem}
\usepackage[dvipsnames]{xcolor}
\usepackage[ruled,noresetcount]{algorithm2e}
\usepackage[pagebackref=false,breaklinks=true,colorlinks,bookmarks=false]{hyperref}
\definecolor{citecolor}{RGB}{34,139,34}

\usepackage{array}
\usepackage{url}

\usepackage{booktabs}
\usepackage{multirow}
\usepackage{threeparttable}
\usepackage{color, colortbl}
\usepackage{changes}
\usepackage{subcaption}
\usepackage{wrapfig}
\usepackage{diagbox}
\usepackage{arydshln} 
\usepackage{makecell}

\usepackage{pifont}

\usepackage{booktabs, makecell, multirow, xcolor}
\definecolor{best}{RGB}{230,230,230}  %
\usepackage{siunitx,booktabs,xcolor,colortbl}
\newcommand\best[1]{\textbf{#1}}
\newcommand\second[1]{\underline{#1}}

\makeatletter
\newcommand{\thickhline}{%
    \noalign {\ifnum 0=`}\fi \hrule height 0.5pt
    \futurelet \reserved@a \@xhline
}
\makeatother
\definecolor{LightGray}{gray}{0.9}

\newcommand{\eg}{\textit{e.g.}}

\begin{document}
%
\title{FreeLong++: Training-Free Long Video Generation via Multi-band SpectralFusion}

%
%
%
%

\newcommand{\zongxin}[1]{{#1}}
\newcommand{\new}[1]{{#1}}

\author{Yu Lu, ~Yi~Yang
\thanks{

%

Y. Lu, and Y. Yang are with ReLER, CCAI, Zhejiang University, Hangzhou, 310027, China (e-mail: \{aniki.yulu, yangyics\}@zju.edu.cn).
}}

\markboth{IEEE TRANSACTIONS ON PATTERN ANALYSIS AND MACHINE INTELLIGENCE}%
{Shell \MakeLowercase{\textit{et al.}}: Bare Demo of IEEEtran.cls for Computer Society Journals}
\IEEEtitleabstractindextext{%
\begin{abstract}

%

%
%
%
%
%
%
Recent advances in video generation models have enabled high-quality short video generation from text prompts. However, extending these models to longer videos remains a significant challenge, primarily due to degraded temporal consistency and visual fidelity. Our preliminary observations show that naively applying short-video generation models to longer sequences leads to noticeable quality degradation. Further analysis identifies a systematic trend where high-frequency components become increasingly distorted as video length grows—an issue we term \textit{high-frequency distortion}.
To address this, we propose \textbf{FreeLong}, a training-free framework designed to balance the frequency distribution of long video features during the denoising process. FreeLong achieves this by blending global low-frequency features, which capture holistic semantics across the full video, with local high-frequency features extracted from short temporal windows to preserve fine details.
Building on this, \textbf{FreeLong++} extends FreeLong’s dual-branch design into a multi-branch architecture with multiple attention branches, each operating at a distinct temporal scale. By arranging multiple window sizes from global to local, FreeLong++ enables multi-band frequency fusion from low to high frequencies, ensuring both semantic continuity and fine-grained motion dynamics across longer video sequences.
Without any additional training, FreeLong++ can be plugged into existing video generation models~(\eg~Wan2.1 and LTX-Video) to produce longer videos with substantially improved temporal consistency and visual fidelity. We demonstrate that our approach outperforms previous methods on longer video generation tasks~(\eg~\textbf{4$\times$}~ and \textbf{8$\times$} of native length). It also supports coherent multi-prompt video generation with smooth scene transitions and enables controllable video generation using long depth or pose sequences. Additional results and details are available on the project website: \url{https://freelongvideo.github.io/}

\end{abstract}

}

\maketitle
\IEEEdisplaynontitleabstractindextext
%
\IEEEpeerreviewmaketitle

\IEEEraisesectionheading{\section{Introduction}\label{sec:introduction}}

\begin{figure*}[tb]
    \centering
    \includegraphics[width=1.0\linewidth]{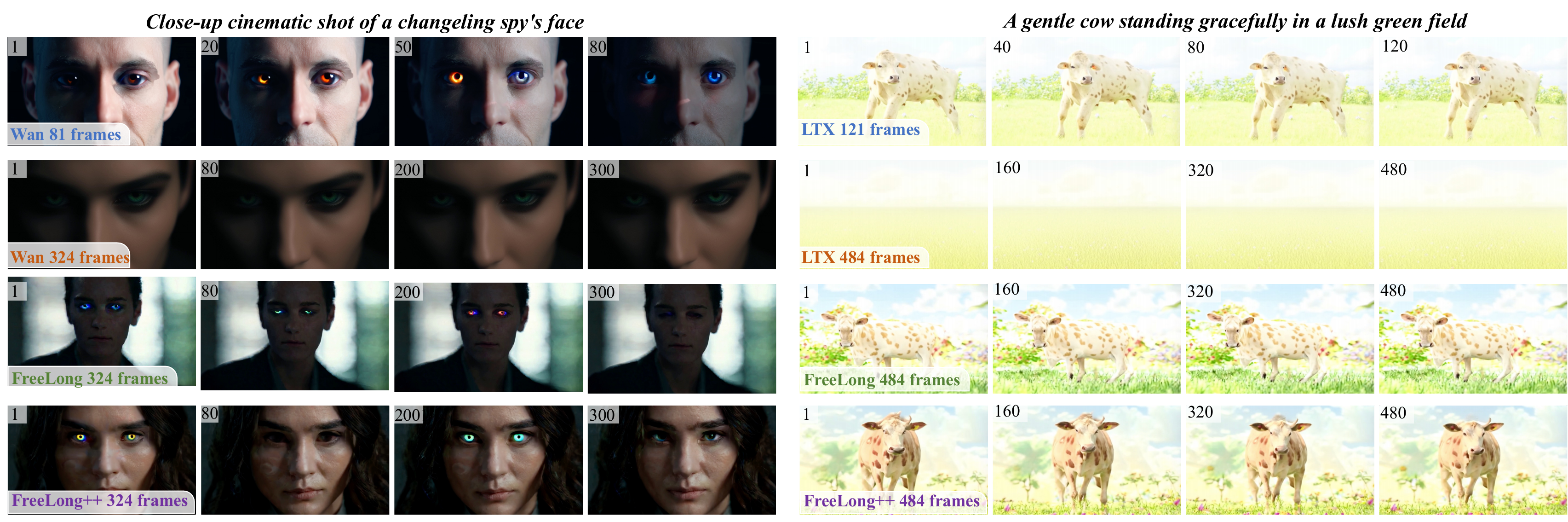}

\vspace{-1.5mm}
    \caption{    
Results of Short and Longer Videos. The first row of each case shows short videos
generated using short video diffusion models (81 frames for Wan-2.1~\cite{wan} and 121 frames for LTX-Video~\cite{ltxvideo}). Directly extending these models to longer videos, like those with 4$\times$~(324 frames and 484 frames), preserves temporal consistency but lacks
fine spatial-temporal details. In contrast, our proposed FreeLong and FreeLong++ adapts short video diffusion models
to create consistent long videos with high fidelity.
    }
    \label{fig:results_comparison}
    
\vspace{-3.5mm}

\end{figure*}
\begin{figure}
\centering
   \setlength{\abovecaptionskip}{0.5cm}
   \includegraphics[scale = 0.4]{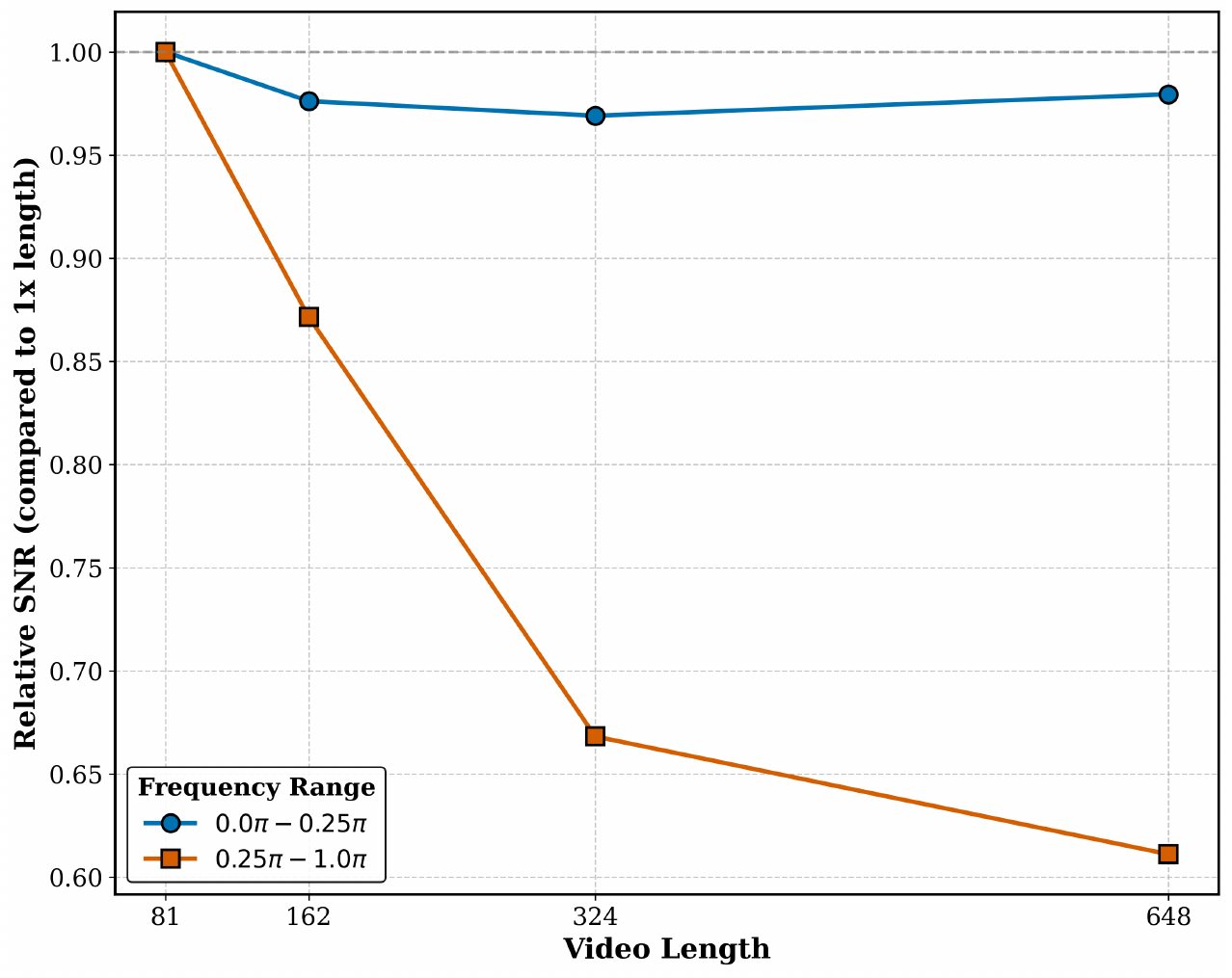}
     \caption{
     \textbf{Ratio of short video SNR on high~(0.25$\pi$-1.0$\pi$)/low~(0.0$\pi$-0.25$\pi$) frequency to longer videos.} 
    Our findings reveal that when direct extend short video diffusion model to generate longer videos, the SNR of high-frequency components in the space-time frequency domain degrades significantly as video length increases.
              }
   \label{fig:frequency_local_global}
\end{figure}

\begin{figure}
\centering
   \setlength{\abovecaptionskip}{0.5cm}
   \includegraphics[scale = 0.45]{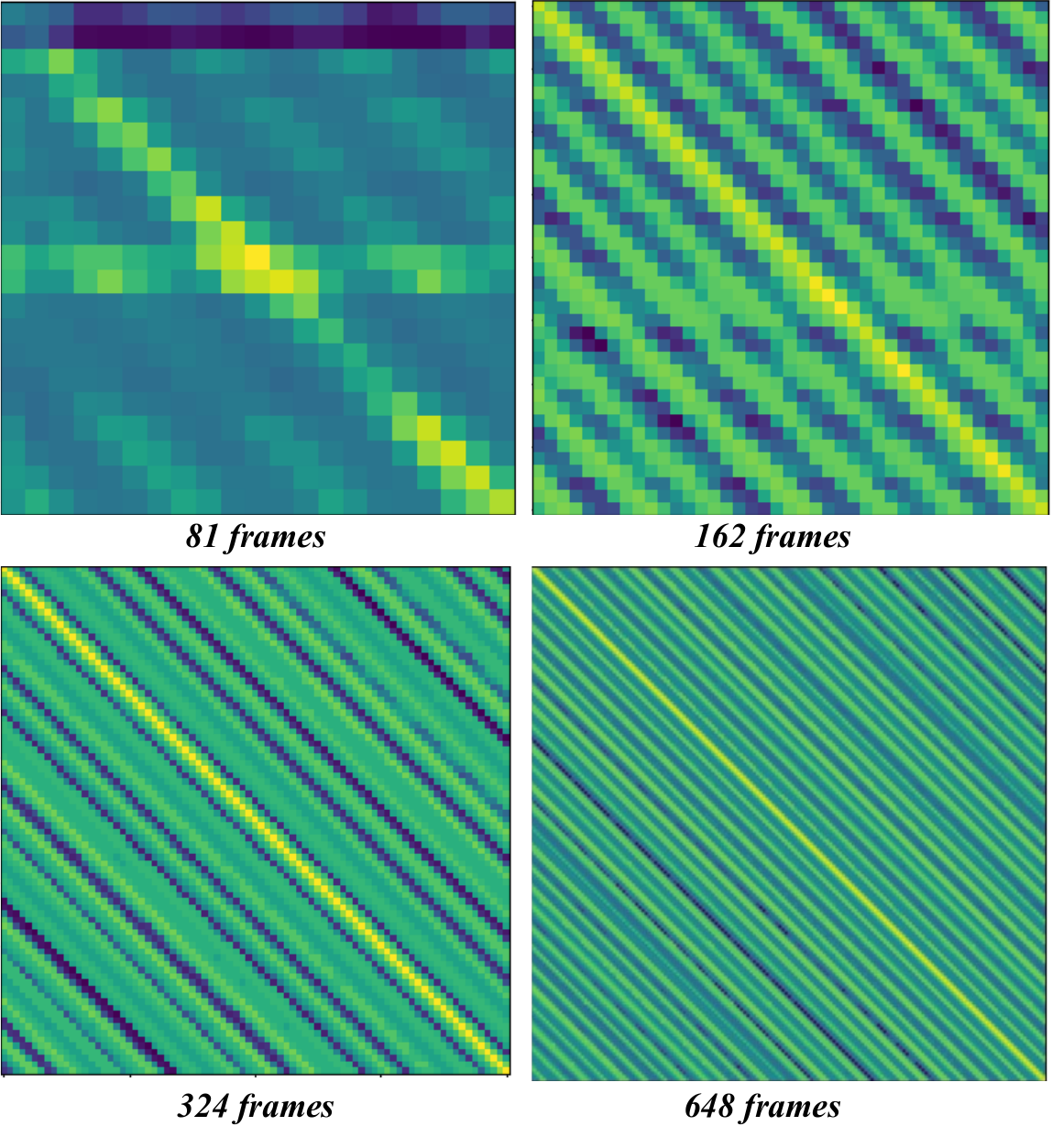}
   \vspace{-0.3cm}
     \caption{
     \textbf{Attention Visualization.} We visualize the attention by average across all layers and time steps from Wan2.1~\cite{wan}. The attention maps for 81-frame videos exhibit a diagonal-like pattern, indicating a high correlation with adjacent frames, which helps preserve high-frequency details and motion patterns when generating new frames. In contrast, attention maps for longer videos are less structured, such as 648 frames~(8$\times$), making the model struggle to identify and attend to the relevant information across distant frames. This lack of structure in the attention maps results in the distortion of high-frequency components of long videos, which results in the degradation of fine spatial-temporal details.
         }
   \label{fig:attention}
\end{figure}

\begin{figure*}[tb]
    \centering
    \includegraphics[width=1.0\linewidth]{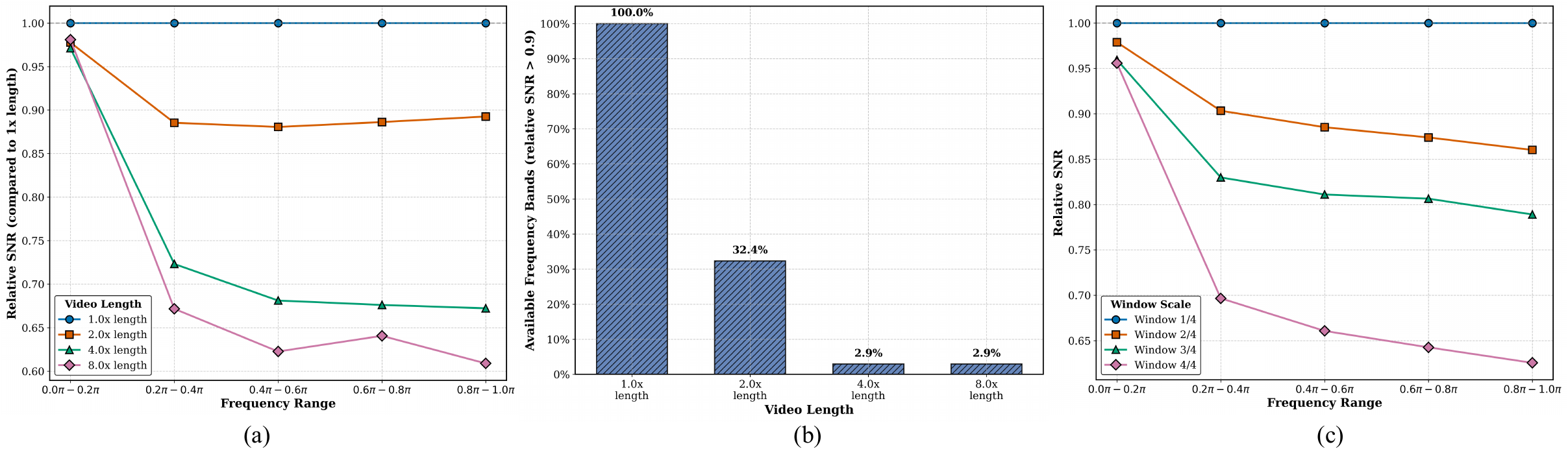}

\vspace{-1.5mm}
    \caption{\textbf{Fine-grained frequency analysis on longer video generation.} 
(a) As video length increases, both the range and severity of frequency distortion grow substantially.
(b) We define available frequency bands as those with a relative SNR above 0.9. As shown, the number of available bands drops significantly when the video length increases from 2$\times$ to 4$\times$, indicating that a fixed two-branch structure in FreeLong is insufficient for modeling motion dynamics in longer sequences.
(c) High-frequency distortion correlates with attention window size: larger window sizes introduce more severe distortion in the high-frequency components.
    }
    \label{fig:freq}
    
\vspace{-3.5mm}

\end{figure*}

Recent advances in video generation models~\cite{videocrafter, cogvideox, animatediff, flowzero, yang2024eva,hunyuanvideo,wan,ltxvideo,stept2v,cosmos,moviegen,pyramidal,open-sora-plan,stablevideo}, have enabled the generation of high-quality short videos from text prompts. These models are typically trained on large-scale video-text datasets~\cite{internvid,webvid,panda70m,koala36m,vidgen1m,hoigen1m,egovid5m}, and their ability to produce coherent short clips has inspired research into extending them to long-form video generation~\cite{streamt2v, vidu, keling, panda70m, hd130m, vlogger,framepack,longcontext,longcontextuning,ttt,videorag,videoauteur,owl1}.
Yet, building long-video generation models requires extensive computational resources and access to large-scale long-video annotations, making them impractical for lightweight and general applications.

A more efficient and practical alternative is to adapt pre-trained short video generation models to generate longer video sequences in a training-free manner. Recent studies~\cite{freenoise,genlvideo,fifo,vstar,riflex,ditctrl,freepca,longdiff,scalingnoise} have explored attention mechanisms~\cite{freenoise,freepca,vstar}, auto-regressive architectures~\cite{fifo,scalingnoise}, and positional encoding~\cite{riflex} to improve long-range consistency in video clips. 
However, these approaches often focus on maintaining coherence at the boundaries of adjacent clips rather than enforcing a unified narrative or consistent visual identity across the entire video. As a result, artifacts such as identity drift, inconsistent lighting, and abrupt scene transitions can emerge, particularly in videos with prolonged durations or complex motion dynamics.

In this study, we propose a straightforward, training-free method to adapt existing short video generation models for generating consistent longer videos. We first evaluate the direct application of short video generators, such as Wan2.1~\cite{wan} (native length 81 frames), to longer sequences (\eg, 4$\times$ video length, 324 frames). As shown in Figure~\ref{fig:results_comparison}, this approach ensures global consistency but results in lower-quality outputs, including blurred textures, and motion jitter beyond the model’s native frame length (see the first and second row of Figure~\ref{fig:results_comparison}).

To understand these issues, we performed frequency analysis on generated long videos. Frequency analysis of generated longer videos revealed stable low-frequency components but \textit{significant distortion in high-frequency components as video length increased} (Figure~\ref{fig:frequency_local_global}). 
In a fine-grained frequency analysis, we also observe increasing distortion in high-frequency components as video length grows (see Figure~\ref{fig:freq}(a)). For example, with double-length sequences, only 30\% of the low-frequency content available, leaving 70\% of high-frequency components distorted; at $4\times$
length, distortion rises to 95\% (Figure~\ref{fig:freq} (b)). This diminishes fine details in longer sequences, such as cat fur or tree leaves becoming blurred (Figure~\ref{fig:results_comparison}, second row).

In this paper, we introduce \textbf{FreeLong}, a novel framework that employs SpectralBlend Attention to balance the frequency distribution of long video features in the denoising process. FreeLong integrates global and local features via two parallel streams, enhancing the fidelity and consistency of long video generation. The global stream deals with the entire video sequence, capturing extensive dependencies and themes for narrative continuity. Meanwhile, the local stream focuses on shorter frame subsequences to retain fine details and
smooth transitions, preserving high-frequency spatial and temporal information. FreeLong combines global and local video features in the frequency domain, improving both consistency and fidelity by blending low-frequency global components with high-frequency local components. 

Building on the FreeLong, we further present \textbf{FreeLong++}, a comprehensive extension of FreeLong that leverages \textit{Multi-band SpectralFusion~(MSF)} framework. Rather than restricting attention to a binary global-local structure, FreeLong++ utilizes multiple attention branches with varying window sizes, where each window attends to a different temporal scale. This design allows us to decompose the video signal into interpretable temporal frequency bands: longer windows capture global semantic continuity and low-frequency structure, while shorter windows focus on fast-changing motion and high-frequency texture. We further propose a multi-band fusion strategy to adaptively merges the multi-window video features in the frequency domain, ensuring that all frequency bands are properly integrated and reconstructed into a consistent video sequence, results in frequency-aligned fusion.

FreeLong++ retains the training-free advantage of FreeLong, introducing no additional model parameters or fine-tuning requirements. Its modular design seamlessly integrates with existing diffusion transformers~\cite{wan,ltxvideo} by directly replacing attention modules in modern video diffusion transformers. Experimental results demonstrate that FreeLong++ significantly outperforms existing training-free baselines by consistently enhancing temporal consistency and visual fidelity, robustly extending short video generation models to generate videos 4 or 8 times longer. Moreover, FreeLong++ effectively supports sophisticated video generation tasks involving complex controls such as pose-guidance or depth-guidance.

Our contributions can be summarized as follows: 
\textbf{1)} We conduct a frequency analysis on the direct application of short video models for longer video generation and identify \textit{high-frequency distortions} in the longer videos. 
\textbf{2)} We propose \textbf{FreeLong} with a SpectralBlend Attention mechanism to merge the consistent low-frequency components of global videos with the high-fidelity high-frequency components of local videos. 
\textbf{3)} We propose \textbf{FreeLong++}, a novel training-free framework built upon, FreeLong. FreeLong++ introduces \textit{Multi-band SpectralFusion (MSF)}, enabling multi-window attention mechanisms to effectively capture temporal dynamics across various frequency bands without additional training or parameters.


\section{Related Work}\label{sec:related_works}

\subsection{Text-to-Video Generation Models}

\noindent Text-to-video (T2V) generation has made significant advancements with the rise of diffusion-based models~\cite{videocrafter, animatediff, flowzero, cogvideox, hunyuanvideo, yang2024eva}, demonstrating remarkable capabilities in generating high-quality, temporally coherent videos. Early video diffusion models leveraged pre-trained image diffusion UNets~\cite{sd} and enhanced them with temporal attention mechanisms to effectively model frame-to-frame dependencies. Notable examples, such as LaVie~\cite{lavie} and VideoCrafter2~\cite{videocrafter}, trained on large-scale video-text datasets like WebVid~\cite{webvid} and InternVid~\cite{internvid}, have been successful in producing high-quality videos of fixed short durations, typically around 2 seconds.

The field has further evolved with the introduction of Sora~\cite{sora}, which highlights the scalability and effectiveness of diffusion transformer~(DiT) architectures~\cite{dit}. Recent innovations, including CogVideoX~\cite{cogvideox}, Mochi1~\cite{mochi}, HunyuanVideo~\cite{hunyuanvideo}, LTX-Video~\cite{ltxvideo}, and Wan2.1~\cite{wan}, have adopted the DiT framework, achieving state-of-the-art performance in video generation. By scaling both model size and the volume of training data, these DiT-based models have managed to extend video generation capabilities to sequences as long as 5 seconds.

Nonetheless, generating longer videos remains a significant challenge. Key bottlenecks include the complexity of temporal modeling, the memory requirements for handling extended video sequences, and the lack of training data annotated for long-range video dependencies. Progress in addressing these limitations is critical to unlocking the potential of T2V systems for generating longer, high-quality videos with enhanced temporal consistency.

\subsection{Long Video Generation}

\noindent Recent efforts~\cite{streamt2v, vidu, genlvideo, panda70m} have explored scaling video diffusion models to longer durations by modifying training objectives or architectures. Approaches such as StreamingT2V~\cite{streamt2v} and Vidu~\cite{vidu} adopt autoregressive generation pipelines or memory-augmented modules to maintain cross-segment consistency. However, these methods are computationally expensive and require extensive retraining on curated long-video datasets.
Additionally, recent autoregressive models~\cite{causvid,framepack,magi-1} fine-tune pretrained short-video diffusion models using a next-clip prediction paradigm. However, such methods are prone to error accumulation during inference, leading to degradation issues such as semantic drift and content forgetting.
To reduce training costs, lightweight alternatives such as Gen-L-Video~\cite{genlvideo} and FreeNoise~\cite{freenoise} introduce training-free extensions based on sliding-window attention and noise rescheduling. While efficient, these approaches suffer from limited temporal modeling capacity and fail to adequately preserve frequency structures, often resulting in temporal drift over extended sequences.
In contrast, we propose FreeLong, a training-free method that enhances longer video generation by blending global low-frequency and local high-frequency features through a dual-branch SpectralBlend Temporal Attention mechanism. Building on this, FreeLong++ introduces a multi-band extension with multiple attention branches of varying window sizes, enabling adaptive modeling across temporal frequency bands and improving consistency and fidelity in longer video sequences.

\section{Preliminary}\label{sec:revisit}

Current video generation models generally adopt a common backbone design to effectively model relationships across spatial and temporal dimensions. Architectures such as UNet~\cite{sd,lavie,videocrafter} and Transformers~\cite{dit,wan} are commonly employed to facilitate the iterative denoising process~\cite{ddim,flowmatch}. The UNet architecture is effective due to its separate spatial and temporal attention layers, which help reduce computational costs, although it may struggle to maintain strong consistency in capturing dependencies.  
Transformer-based models are effective at modeling long-range dependencies in data by using 3D attention mechanisms. These mechanisms capture both spatial and temporal relationships, making the models well-suited for complex video sequences. The attention mechanism used in both UNet and transformers is defined as:

\[
\mathbf{A} = \text{Softmax}\left(\frac{\mathbf{Q} \mathbf{K}^T}{\sqrt{d_k}}\right) \mathbf{V},
\]

where \( \mathbf{Q} \), \( \mathbf{K} \), and \( \mathbf{V} \) are the query, key, and value matrices, and \( d_k \) is the key dimensionality. This mechanism can be applied to spatial, temporal, or spatiotemporal dimensions.

Additionally, control signals such as text, depth, or pose can be seamlessly incorporated by modifying \( \mathbf{K} \) and \( \mathbf{V} \)to include the relevant control features. This enables the generation of contextually guided and semantically rich video content.

While these advancements allow for the generation of coherent and high-quality video frames, generating longer video sequences remains a significant challenge. Video generation models, generally pretrained on shorter videos, often struggle with maintaining consistent quality over longer sequences. The attention mechanisms, though powerful, tend to degrade in effectiveness when tasked with modeling long-range dependencies, ultimately leading to reduced video quality as the sequence length increases.

\section{Methodology}\label{sec:method}

In this section, we first introduce \textbf{FreeLong}, which adopts a two-branch SpectralBlend strategy to fuse global low-frequency context with local high-frequency details, thereby maintaining semantic continuity and visual fidelity. 
Building on this, we introduce \textbf{FreeLong++}, extending SpectralBlend to a multi-branch approach with finer frequency band control for enhanced motion dynamics.

\subsection{FreeLong}\label{sec:freelong}

\subsubsection{Observation and Analysis}

\noindent When attempting to adapt short video diffusion models to generate longer videos, a straightforward approach is to input a longer noise sequence into the short video models. The transformer attention layers in the video generation model are not constrained by input length, making this method seemingly viable. However, our empirical study reveals significant challenges, as demonstrated in Figure~\ref{fig:results_comparison}. Generated longer videos often exhibit fewer detailed textures, such as blurred fur in the cat, and more irregular variations, like abrupt changes in motion. We attribute these issues to two main factors: the limitations of the attention mechanism and the distortion of high-frequency components.

\noindent\textbf{Attention Mechanism Limitations:} The attention mechanism in video generation models, pre-trained on short videos, struggles to generate longer videos effectively. As shown in Figure~\ref{fig:attention}, for a DiT model trained on 81-frame videos, attention maps exhibit a clear diagonal pattern, reflecting strong correlations between adjacent frames and preserving spatial-temporal details and motion patterns. However, with 324-frame videos~(4$\times$) or 648 frame videos~(8$\times$), the attention maps lose structure, making it harder to capture relevant information over distant frames. This results in missed subtle motion patterns and over-smoothed or blurred outputs.

\noindent\textbf{Frequency Analysis:} To better understand the generation process of long videos, we analyzed the frequency components in videos of varying lengths using the Signal-to-Noise Ratio (SNR) as a metric. Ideally, short video diffusion models generate short videos with high quality. Robust longer videos, such as $4 \times$ the original length derived from such models, should exhibit consistent SNR values across all frequency components. However, Figure~\ref{fig:frequency_local_global} reveals significant differences in the SNR of high/low frequency components\footnote{We split the frequency components into high-frequency (\( \phi \sim (0.25\pi - 1.00\pi) \)) and low-frequency (\( \phi \sim (0.00\pi - 0.25\pi) \)) and compared the SNR of each component in longer videos to the corresponding SNR in short videos.} between generated short and longer videos. The SNR of low-frequency components remains relatively consistent for long videos (1.0 for origin length frames to 0.97 for $8 \times$ frames), suggesting that the model maintains overall structure and low-frequency details in extended sequences. However, the SNR of high-frequency components drops significantly for longer videos (1.0 for origin length to 0.6 for $8 \times$ length), indicating a loss of fine details and increased distortion, leading to suboptimal visual fidelity.

\begin{figure*}[tb]
    \centering
    \includegraphics[width=1.0\linewidth]{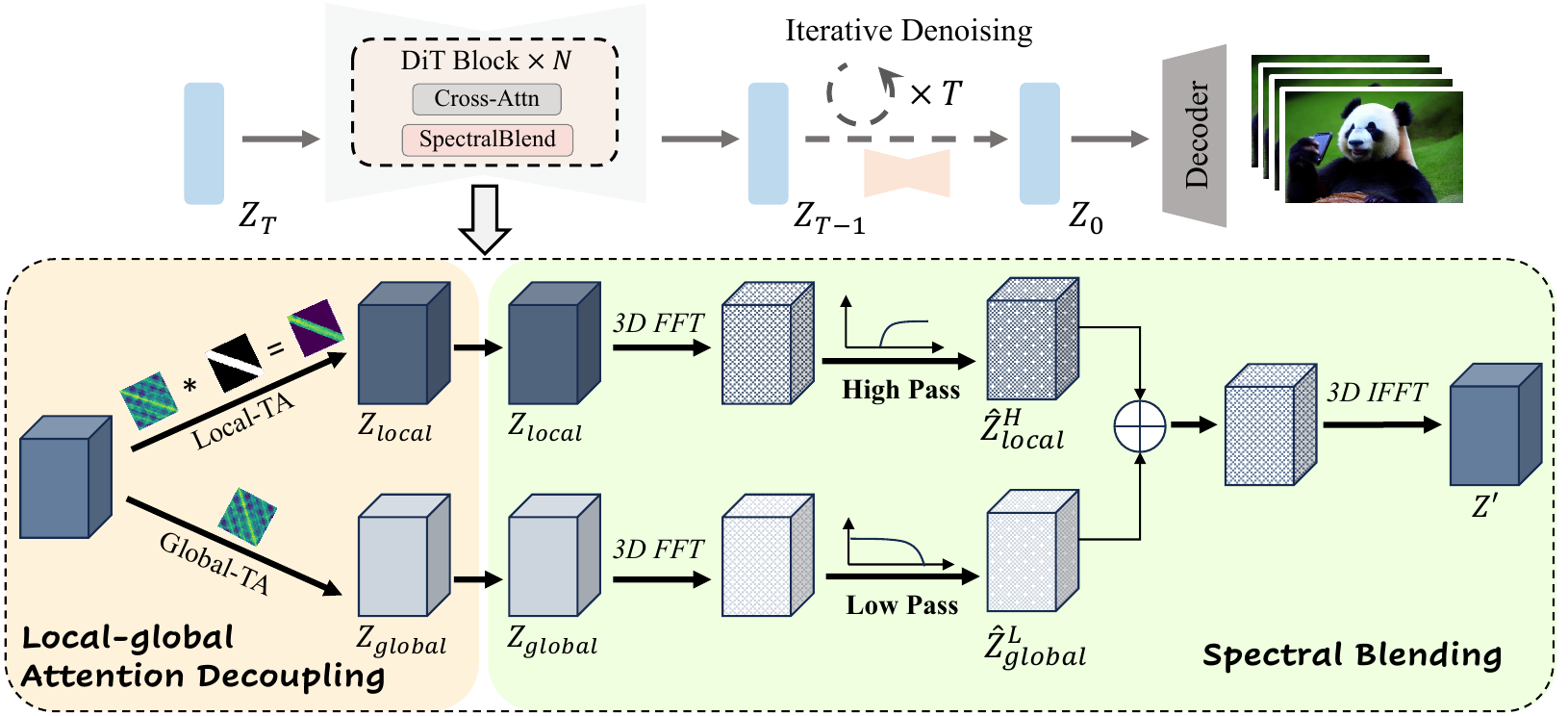}

\vspace{-1.5mm}
    \caption{    \textbf{Overview of FreeLong.} FreeLong facilitates consistent and high-fidelity video generation using SpectralBlend Attention. SpectralBlend effectively blends low-frequency global video features with high-frequency local video features through a two-step process: local-global attention decoupling and spectral blending. Local video features are obtained by masking temporal attention to concentrate on fixed-length adjacent frames, while global temporal attention encompasses all frames. During spectral blending, 3D FFT projects features into the frequency domain, where high-frequency local components and low-frequency global components are merged. The resulting blended feature, transformed back to the time domain via IFFT, is then utilized in the subsequent block for refined video generation.
    }
    \label{fig:freelong_framework}
    
\vspace{-3.5mm}

\end{figure*}
Motivated by the frequency analysis, we propose FreeLong, a method designed to generate high-fidelity and consistent long videos using the inherent power of the diffusion model. As illustrated in Figure~\ref{fig:freelong_framework}, our FreeLong uses a pre-trained short video generation models and introduces a SpectralBlend attention to facilitate long video generation. The SpectralBlend attention consists of two steps: local-global attention decoupling and spectral blending.

\subsubsection{Local-global Attention Decoupling}
The attention in short video models is optimized to model short frame sequences accurately, maintaining high-fidelity visual information. 
Conversely, the long-range attention from short video models tends to maintain overall layout and and object consistency.  
Given these properties, we first decouple the local and global attention. For a video sequence with length $T$, let $i$ and $j$ denote the indices of query and key frames, respectively.The local attention matrix can be obtained as:
\begin{equation}
A_{\text{local}}(i, j) = 
\begin{cases}
\text{Softmax}\left(\dfrac{Q_i K_j^\top}{\sqrt{d}}\right) & \text{if } |i - j| < \left\lfloor \dfrac{T_\alpha}{2} \right\rfloor \\
0 & \text{otherwise},\nonumber
\end{cases}
\end{equation}
where \( Q \) and \( K \) are the query and key matrices derived from the input video feature~\( Z_{in} \). The local attention \( A_{\text{local}} \) leads to each frame \( i \) only attending to frames within a window of \( T_\alpha \) frames. We set \( T_\alpha \) as the native video length of pretrained models~(\eg, 81 frame for Wan2.1~\cite{wan}).
Given the local attention matrix \( A_{\text{local}} \), the local video features \( Z_{\text{local}} \) can be obtained by:
$Z_{\text{local}} = A_{\text{local}} V$, 
where \( V \) is the value matrix derived from the input video feature \( Z_{in} \).
By restricting the attention to adjacent local frames, we preserve the capabilities of short video models, thereby retaining high-fidelity visual details in local video features.

We then define the global attention matrix where each frame attends to all other frames. The global attention matrix can be computed as follows:
\begin{equation}
A_{\text{global}}(i, j) = \text{Softmax}\left(\frac{Q_i K_j^\top}{\sqrt{d}}\right). \nonumber
\end{equation}
Given the global attention matrix \( A_{\text{global}} \), the global video features \( Z_{\text{global}} \) can be obtained by:
$Z_{\text{global}} = A_{\text{global}} V$. 
The global video features process the entire video sequence, ensuring narrative continuity and consistency, while capturing long-range dependencies and overarching themes.

\subsubsection{Spectral Blending}  
After obtaining the global and local video features, a frequency filter is used to blend the low-frequency components of the global video latent \( Z_{global} \) with the high-frequency components of the local video latent \( Z_{local} \), resulting in a new video latent \( Z' \). This fused latent retains the global consistency and structure provided by \( Z_{global} \), while benefiting from the enhanced high-frequency details introduced by \( Z_{local} \). The process is described by:
\begin{align}
& \hat{Z}^L_{{global}} = \mathcal{F}_\text{3D}(Z_{global}) \odot \mathcal{P},\nonumber  \\
& \hat{Z}^H_{{local}} = \mathcal{F}_\text{3D}(Z_{local}) \odot (1 - \mathcal{P}),\nonumber \\
& Z' = \mathcal{F}^{-1}_\text{3D}(\hat{Z}^L_{{global}} + \hat{Z}^H_{{local}}),\nonumber
\end{align}
where \(\mathcal{F}_\text{3D}\) is the Fast Fourier Transformation operated on both spatial and temporal dimensions, \(\mathcal{F}^{-1}_\text{3D}\) is the Inverse Fast Fourier Transformation that maps back the blended representation \( Z' \) from the frequency domain, and \(\mathcal{P} \in \mathbb{R}^{4 \times N \times h \times w}\) is the spatial-temporal Low Pass Filter (LPF), which is a tensor of the same shape as the latent. The final fused video feature \( Z' \) serves as the input to our subsequent video generation module. 

The rationale behind using low-frequency components from the global video features and high-frequency components from the local video features stems from our analysis. The global features provide a stable, consistent structure, preserving the overall layout and object consistency throughout the video. This is crucial for maintaining temporal consistency in long videos. On the other hand, local features retain high-fidelity details, which are essential for capturing fine textures and intricate motion patterns that tend to degrade in long sequences. By blending these components in the frequency domain, we harness the strengths of both global consistency and local detail preservation, addressing the issues of blurred frames and temporal flickering observed in our analysis.

\subsubsection{Implementation details}
We apply FreeLong on state-of-the-art diffusion transformer models, Wan-2.1~\cite{wan} and LTX-Video~\cite{ltxvideo}. Wan models can generate high-quality 81 frames/5s videos, and LTX-Video can generate 121 frame videos. 
We set $T_\alpha$ same with native video length for the local attention
setting. During inference, the parameters of the frequency filter for each model are kept the same for a fair comparison. Specifically, we use a Gaussian Low Pass Filter with a normalized spatiotemporal stop frequency of D$_{0}$ = 0.25.

\subsection{FreeLong++}\label{sec:freelong++}

\subsubsection{Observation}\label{sec:freelong++_observation}

As discussed previously, \textbf{FreeLong} uses a dual-branch SpectralBlend attention mechanism to separately model global low-frequency context and local high-frequency details. While this two-branch architecture is effective for moderately extended video sequences, it encounters significant limitations as video length increases, most notably in the form of increased frequency distortion. As illustrated in Figure~\ref{fig:freq}(a), increasing the video length results in a pronounced degradation of high-frequency components, with both the severity and the range of affected frequencies growing substantially. To quantify this effect, we define ``distorted” frequency bands as those with a relative signal-to-noise ratio (SNR) below 0.9. Our analysis shows that the proportion of such distorted bands increases sharply with extended video durations. For example, at four times the native video length $T_{\alpha}$, only about 3\% of the frequency bands remain reliable (Figure~\ref{fig:freq}(b)). This dramatic decline in high-frequency fidelity underscores the inadequacy of the simple dual-branch approach in handling the complex shifts in frequency distributions inherent to long sequences, highlighting the need for an adaptive, more refined frequency decomposition strategy. 

Furthermore, our experiments show that adjusting the temporal attention window size significantly influences high-frequency distortion patterns. As depicted in Figure~\ref{fig:freq}(c), when the generated video length is fixed at $4\times T_{\alpha}$ (four times the native video length $T_{\alpha}$), varying the temporal attention window size yields distinct patterns of frequency degradation.
This observation directly motivated the design of FreeLong++, which employs a multi-branch attention architecture to provide finer-grained control at different temporal scales. This design significantly enhances the model’s ability to preserve long-range consistency and accurately capture complex motion dynamics. 

\subsubsection{Overview}\label{sec:freelong++_overview}
\begin{figure*}[tb]
    \centering
    \includegraphics[width=1.0\linewidth]{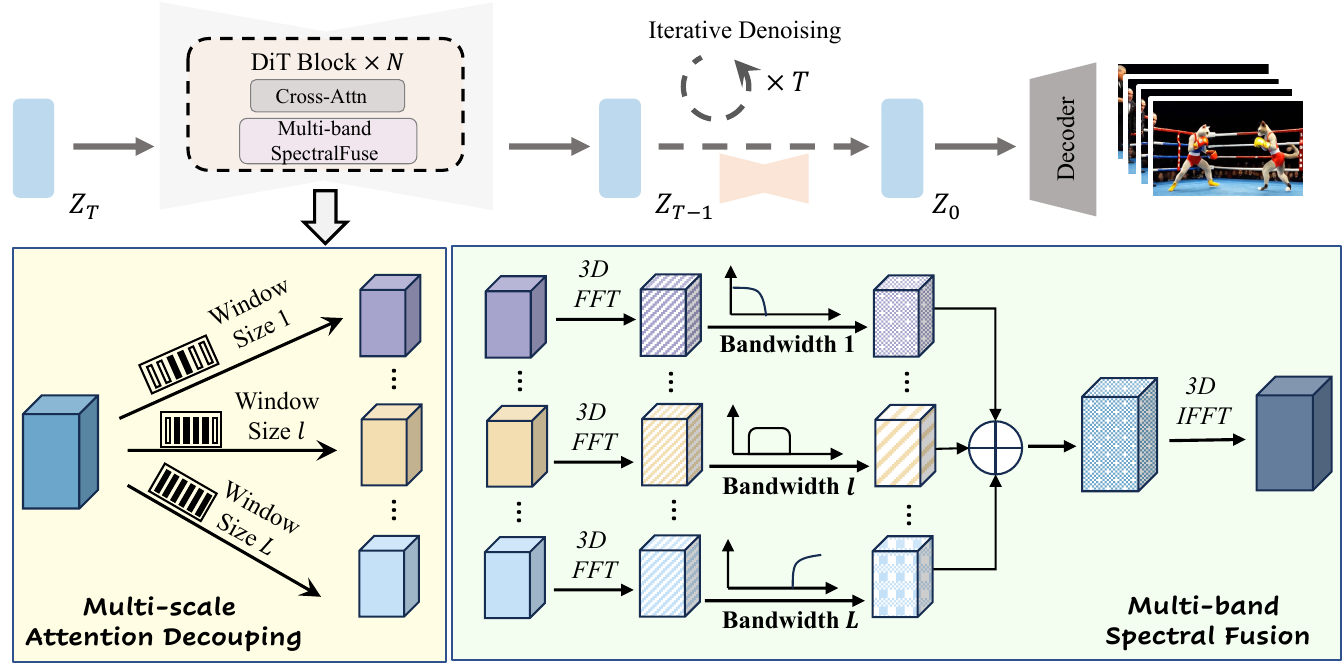}

\vspace{-1.5mm}
    \caption{    \textbf{Overview of FreeLong++.} The FreeLong++ framework extends FreeLong by introducing Multi-band SpectralFusion Attention. Multi-scale temporal branches with varying window sizes capture motion dynamics at different frequency bands. Each branch is processed in the frequency domain and selectively fused via scale-specific filters, enhancing long-range consistency while preserving fine-grained motion.
             }
    
    \label{fig:freelong++_framework}
    
\vspace{-3.5mm}

\end{figure*}

Guided by these insights, we propose FreeLong++, whose framework is illustrated in Figure~\ref{fig:freelong++_framework}. Leveraging a diffusion transformer architecture with integrated 3D attention mechanisms across spatial and temporal dimensions, FreeLong++ incorporates multiple attention branches designed to effectively capture dynamics at varying temporal scales.

Specifically, we extend the spectral blending mechanism into a multi-branch attention architecture, where each branch independently focuses on a distinct temporal scale. These scales range from short-term branches (capturing immediate local spatial-temporal features), through mid-term branches (capturing intermediate-level motion patterns and dependencies), to long-term branches (aggregating comprehensive global temporal contexts). Each attention branch employs a dedicated frequency-domain band-pass filter, enabling selective extraction and emphasis of frequency-specific features pertinent to its temporal scope. The outputs from these branches are subsequently combined in the frequency domain, producing a composite representation that effectively integrates short-term dynamic details with broader, long-term structural consistency.

\subsubsection{Multi-Scale Attention Decoupling}
To capture dynamics at different temporal ranges, we decouple the original temporal attention into multiple parallel scale-specific attention branches. Each branch $l$ operates on a different temporal window size $\alpha_l T_\alpha$, expressed as a multiple of the native video length $T_\alpha$. For example, a three-scale configuration could use $\alpha_1=1$, $\alpha_2=2$, and $\alpha_3=4$, corresponding to attention windows of length $1\times T_\alpha$, $2\times T_\alpha$, and $4\times T_\alpha$, respectively. 
For a video sequence with length $T$ , let $i$ and $j$ denote the indices of query and key frames, respectively. For each scale $l$ we apply a masked self-attention that limits each query frame to attend only to an interval of $\alpha_l T_\alpha$ frames around it. We denote the resulting masked attention matrix for scale $l$ as

\begin{equation}
A_{_l}(i, j) = \begin{cases} 
\text{Softmax}\left(\frac{Q_i K_j^\top}{\sqrt{d}}\right) & \text{if } |i - j| < \left\lfloor \dfrac{\alpha_l T_\alpha}{2} \right\rfloor, \\
0 & \text{otherwise},\nonumber
\end{cases}
\end{equation}
where $Q$, $K$ are the query and key matrices of the video features. This ensures that branch $l$’s attention is confined to a temporal span of $\alpha_l T_\alpha$ frames. Using this decoupling, we obtain a set of multi-scale video features ${Z_{(l)}}$: the finest-scale branch (small $\alpha_l$) focuses on short-range interactions and preserves high-frequency details, while coarser-scale branches (large $\alpha_l$ up to the full sequence) capture longer-range dependencies and global context (low-frequency structure).

\noindent\textbf{Efficient Attention via Sparse Key Frames:}
To maintain computational efficiency, particularly for the largest temporal window, FreeLong++ propose \textit{sparse attention} through key-frame selection. 
The motivation comes from that long-range temporal relationships often exhibit redundancy and only require a subset of key frames to effectively capture global context~\cite{spargeattn,flashvideo,tileattn,sparse}.
Attention computations in the global-scale~(largest \(\alpha_l\)) branch are restricted to a uniformly sampled subset of representative frames, denoted \(\mathcal{K}\). Formally, the sparsified attention matrix for this largest scale is:

\begin{equation}
A_{\text{sparse}}(i, j) = \begin{cases}
\text{Softmax}\left(\frac{Q_i K_j^\top}{\sqrt{d}}\right) & \text{if } j \in \mathcal{K}, \\
0 & \text{otherwise}.\nonumber
\end{cases}
\end{equation}
The results for most global-branch can be easily obtained by $Z_{\text{sparse}} = A_{\text{sparse}} V_{\text{sparse}}$. 
This strategic sparsification significantly reduces computational overhead while preserving the critical global temporal context necessary for long-range consistency.

\subsubsection{Multi-band Spectral Fusion}
Given the multi-scale features $Z_{(l)}$, we integrate them in the frequency domain to exploit their complementary bandwidths. We first transform each scale’s features into the spectral domain using a 3D Fast Fourier Transform (FFT) over spatial and temporal dimensions.

Formally, let \(Z_l\) denote the latent video features from the \(l\)-th attention branch (with \(l=1\) as the most local branch and \(l=L\) the most global). We project each branch's output into the frequency domain and apply a scale-specific spectral filter before fusing. The multi-band fusion process is described by:

\begin{align}
    \hat{Z}_l &= \mathcal{F}_\text{3D}(Z_l), \quad l = 1,2,\ldots,L,\nonumber \\
    \hat{Z}' &= \sum_{l=1}^{L} \mathcal{P}_l \odot \hat{Z}_l,\nonumber \\
    Z' &= \mathcal{F}^{-1}_\text{3D}(\hat{Z}').\nonumber
\end{align}
Here, $\mathcal{F}_\text{3D}$ and $\mathcal{F}^{-1}_\text{3D}$ denote the 3D Fast Fourier Transform and its inverse, applied over the spatio-temporal dimensions of the latent feature $Z_l$. Each $\hat{Z}_l$ represents the frequency-domain representation of branch $l$'s attention output. The term $\mathcal{P}_l$ is a \textbf{scale-specific frequency mask} (i.e., a band-pass filter), which selectively retains the frequency band corresponding to the temporal scale $\alpha_l$ of branch $l$.

The temporal window $\alpha_l T_\alpha$ for branch $l$ determines its maximum frequency $\frac{1}{2\alpha_l}\pi$ based on the Nyquist criterion\footnote{The Nyquist–Shannon theorem states that a signal whose highest frequency is $f_{\max}$ can be reconstructed only if the sampling rate exceeds $2f_{\max}$; otherwise aliasing occurs.}~\cite{nyquist1,nyquist2}.
For example, the coarsest scale ($\alpha_l = 4$) retains frequencies within $[0, \frac{1}{8}\pi]$, capturing slow, global dynamics. A medium scale ($\alpha_l = 2$) selects $[\frac{1}{8}\pi, \frac{1}{4}\pi]$, while the finest scale ($\alpha_l = 1$) covers the high-frequency range $[\frac{1}{4}\pi, 1.0\pi]$, encoding fast, local motion details.

After filtering, the masked frequency components across all branches are summed to form $\hat{Z}'$, which is then transformed back to the time domain using inverse FFT to produce the final fused latent $Z'$.

The rationale for multi-band spectral fusion is to capture a richer spectrum of motion dynamics while maintaining long-range consistency. In FreeLong++, low-frequency global features (\(Z_1\)) still provide a stable backbone for overall scene structure and temporal consistency across the entire sequence, as in the two-branch case. However, by adding intermediate-scale branches (\(Z_2,\ldots,Z_{L-1}\)), the framework also preserves mid-range dynamics that a single local branch might miss. Each scale-specific filter \(\mathcal{P}_l\) injects the appropriate level of detail: slower temporal changes (e.g., gradual movements or scene transitions) are handled by lower-frequency components, whereas faster motions and fine textures are reinforced by higher-frequency components. The multi-band fusion thus balances the frequency content across scales, preventing both the loss of fine details and the distortion of medium-speed motions. As a result, the fused latent \(Z'\) contains multi-scale temporal information, leading to improved motion realism and smoother transitions.

\subsubsection{SpecMix Noise Initialization}
To stabilize long-range consistency while preserving local details, we introduce SpecMix, an adaptive spectral-domain noise initialization integrated within FreeLong++. SpecMix are based on two critical observations: (i) consistent low-frequency initialization enables models to better synthesize high-frequency details~\cite{wu2023freeinit}, whereas (ii) fully independent noise reduces temporal consistency~\cite{freenoise}. Specifically, we define two noise components: a consistency baseline $x_{\text{base}}$ and a per-frame residual  $x_{\text{res}}$. To construct  $x_{\text{base}}$, we use a sliding-window shuffling procedure inspired by prior work~\cite{freenoise}, where noise segments are shuffled across neighboring temporal windows to enforce consistent low-frequency content. Concurrently, we sample  $x_{\text{res}}$ independently as Gaussian noise, providing controlled local variations.

Both $x_{\text{base}}$ and $x_{\text{res}}$ are then transformed into the spectral domain. We apply a 3D Fast Fourier Transform, yielding frequency-domain tensors:
$x_{\text{base}}^{\mathcal{F}}\!=\mathcal{F}_\text{3D}(x_{\text{base}})$ and  
$x_{\text{res}}^{\mathcal{F}}\!=\!\mathcal{F}_\text{3D}(x_{\text{res}})$.  
For each time index $t$, we compute a normalised distance to the sequence centre,
\[
d_t=\frac{|\,t-(T-1)/2\,|}{(T-1)/2}\in[0,1],
\]
and map it to a mixing angle $\theta_t=d_t\cdot\frac{\pi}{2}$.  
The final spectral representation is then
\[
\tilde{x}^{\mathcal{F}}_t=\cos\theta_t\;x_{\text{base},t}^{\mathcal{F}}
\;+\;
\sin\theta_t\;x_{\text{res},t}^{\mathcal{F}},
\]
where $x_{\text{base},t}^{\mathcal{F}}$ and $x_{\text{res},t}^{\mathcal{F}}$ denote the spectral slices at frame $t$. This formulation ensures that low-frequency (with small $d_t$) rely predominantly on the consistency base noise, while high-frequency (with $d_t$ close to 1) incorporate a larger proportion of the stochastic residual noise. Finally, a 3D inverse FFT are applied to $\tilde{x}^{\mathcal{F}}$ to return to the spatial domain, yielding the initial noise tensor $x_0$ for the diffusion process. Notably, this linear combination~\cite{freebloom} preserves the overall all variance of the magnitude spectra at each temporal slice.

\subsubsection{Implementation details}
We apply FreeLong++ to state-of-the-art diffusion transformer models, Wan-2.1-1.3B~\cite{wan} and LTX-Video~\cite{ltxvideo}. The Wan model generates 81-frame/5s videos, while LTX-Video produces 121-frame videos. For 4$\times$ longer video generation, we use 3 branches with $\alpha_l$ = {1, 2, 4}, and for 8$\times$ longer generation, we use 4 branches with $\alpha_l$ = {1, 2, 4, 8}. Different branch with varing window size can be achieved by simply adjusting the window size in existing attention tools like flash-attention~\cite{flashattn}. We uniformly sample half of the frames as keys in the sparse attention for the global branch.




\section{Experiments}

\subsection{Evaluation Benchmark Details}

\textbf{Test Prompts:} We evaluated our method using 100 augmented prompts randomly selected from VBench-Long~\cite{vbench}.

\noindent\textbf{Evaluation Metrics:} For text-to-video generation, we utilized VBench-Long~\cite{vbench} metrics to assess video consistency and fidelity in long videos.

1. \textbf{Video Consistency:}  
   Subject consistency: Assessed using DINO~\cite{dino} feature similarity across frames to ensure consistent object appearance.  
   Background consistency: Measured using CLIP~\cite{clip} feature similarity across frames.  
   Motion smoothness: Evaluated using motion priors in the AMT~\cite{amt} video frame interpolation model.

2. \textbf{Video Fidelity:}  
   Temporal flickering: Determined by computing mean absolute differences across static frames.  
   Image quality: Measured using the MUSIQ~\cite{musiq} image quality predictor trained on the SPAQ~\cite{spaq} dataset.
   Aesthetic Quality: We evaluate the artistic and beauty value perceived by humans towards each video frame using the LAION aesthetic predictor~\cite{laion_aesthetic_predictor}

For faster experiments, we generate videos 4$\times$ longer for each base model~(Wan-1.3B~\cite{wan} and LTX-Video~\cite{ltxvideo}) in the ablation study and also provide 8$\times$ longer video generation in our experiments.
For controllable long video generation, such as pose- or depth-guided videos, we utilized VACE~\cite{vace} as the base model and applied our attention mechanism.

\subsection{Quantitative Comparison}


\begin{table*}[t]
  \centering
  \small
  \setlength{\tabcolsep}{8pt}
  \renewcommand{\arraystretch}{1.25}
  \caption{\textbf{Quantitative comparison} on the Wan~\cite{wan} model ($4\times$ frames). “Direct sampling” and “Sliding window” indicate directly
sampling 324 frames and applying temporal sliding windows based on short video generation models,
respectively. Compared to these methods, our FreeLong++ achieves consistent long video generation
with high fidelity.
           All scores $\uparrow$.}
  \label{tab:compare_wan}

  \rowcolors{2}{white}{gray!3}
  \begin{tabular}{
      l
      S S S S S S
  }
    \toprule
    \rowcolor{gray!15}
    \textbf{Model} &
    {Subj.\ Cons.} &
    {Back.\ Cons.} &
    {Motion Smooth.} &
    {Temp.\ Flicker} &
    {Imaging Qual.} &
    {Aesthetic Qual.} \\

    \midrule
    Direct sampling      & \second{98.10}             & \second{97.35}             & \second{98.90} & \best{98.88} & 60.52 & 59.07 \\
    Sliding window       & 94.64             & 94.75            & 98.46        & 96.52        & 66.71             & 61.26 \\
    FreeNoise~\cite{freenoise} & 96.05             & 96.31    & 98.06        & 97.63        & \second{67.00}    & 62.35 \\[2pt]
     CausVid~\cite{causvid} & 97.59 & 96.03 & 98.03 & 96.97 & 65.72 & 58.87 \\
    \arrayrulecolor{gray!50}\specialrule{\heavyrulewidth}{0pt}{0pt}
    \rowcolor{gray!3}
    \textit{FreeLong} & 97.85 & 96.85 & 98.92 & 98.29 & 66.33 & \second{62.42} \\   
    \rowcolor{gray!3}
    \textit{FreeLong++} & \best{98.70} & \best{97.83} & \best{98.99} & \second{98.57} & \best{68.82} & \best{64.93} \\

    \bottomrule
  \end{tabular}
\end{table*}

\begin{table*}[t]
  \centering
  \small
  \setlength{\tabcolsep}{8pt}
  \renewcommand{\arraystretch}{1.25}
  \caption{\textbf{Quantitative comparison} on the LTX-Video~\cite{ltxvideo} model ($4\times$ frames).
           All scores $\uparrow$.}
  \label{tab:compare_ltx}

  \rowcolors{2}{white}{gray!3}
  \begin{tabular}{
      l
      S S S S S S
  }
    \toprule
    \rowcolor{gray!15}
    \textbf{Method} &
    {Subj.\ Cons.} &
    {Back.\ Cons.} &
    {Motion Smooth.} &
    {Temp.\ Flicker} &
    {Imaging Qual.} &
    {Aesthetic Qual.} \\

    \midrule
    Direct sampling       & 97.75 & \second{97.57} & \best{99.48} & \second{99.40} & 40.05 & 43.68 \\
    Sliding window        & 96.27 & 96.23 & 99.22 & 90.02 & 46.70 & 49.63 \\
    FreeNoise~\cite{freenoise} & 96.29 & 96.25 & 99.22 & 99.03 & 45.70 & 49.67 \\[2pt]
    \arrayrulecolor{gray!50}\specialrule{\heavyrulewidth}{0pt}{0pt}
    \rowcolor{gray!3}

    \textit{FreeLong}     & \second{98.98} & 97.42 & \second{99.47} & \best{99.40} & \second{45.95} & \second{51.92} \\
    \rowcolor{gray!3}
    \textit{FreeLong++}   & \best{99.55} & \best{97.94} & 99.19 & 99.07 & \best{61.12} & \best{54.68} \\

    \bottomrule
  \end{tabular}
\end{table*}

\begin{table*}[t]
  \centering
  \small
  \setlength{\tabcolsep}{8pt}
  \renewcommand{\arraystretch}{1.25}

\caption{\textbf{Ablation study on each module in Freelong++}. Addition refers to directly summing the outputs of the global and local branches.}
\label{tab:ablation}

  \rowcolors{2}{white}{gray!3}          
  \begin{tabular}{
      l                                    
      S S S S S                            
      >{\raggedleft\arraybackslash}p{18mm} 
  }
    \toprule
    \rowcolor{gray!15}
    \textbf{Method} &
    {Subj.\ Cons.} &
    {Back.\ Cons.} &
    {Motion Smooth.} &
    {Aesthetic Qual.} &
    {Imaging Qual.} &
    \makecell[c]{Infer.\\Time\,($\downarrow$)} \\

    \midrule
    Global-branch & 98.10             & 97.35     & 98.90 & 60.52 & 59.07 & 50\,s \\
    Local-branch       & 95.21 & 95.43 & 97.97 & 66.68 & 61.32 & 22\,s   \\[2pt]

    \arrayrulecolor{gray!50}\specialrule{\heavyrulewidth}{0pt}{0pt}   
    Addition           &   97.18   &   96.40   &  98.85     &   61.47    &   58.64    & 61\,s   \\

    FreeLong   & 97.85 & 96.85 & 98.92 & 66.33 & 62.41 & 72\,s \\
    FreeLong+$SpecMix$   & \best{98.88} & \best{98.25} & \best{99.09} & \second{67.78} & \second{64.40} & 72\,s \\

    FreeLong++         & \second{98.70} & \second{97.83} & \second{98.99} & \best{68.82} & \best{64.93} & 96\,s \\
    FreeLong++$_\text{sparse}$ & 98.60  & 97.73 & 98.98  & 68.65   & 64.52  & 74\,s \\
    \bottomrule
  \end{tabular}
\end{table*}

\begin{figure*}
\centering
   \setlength{\abovecaptionskip}{0.5cm}
   \includegraphics[scale = 0.5]{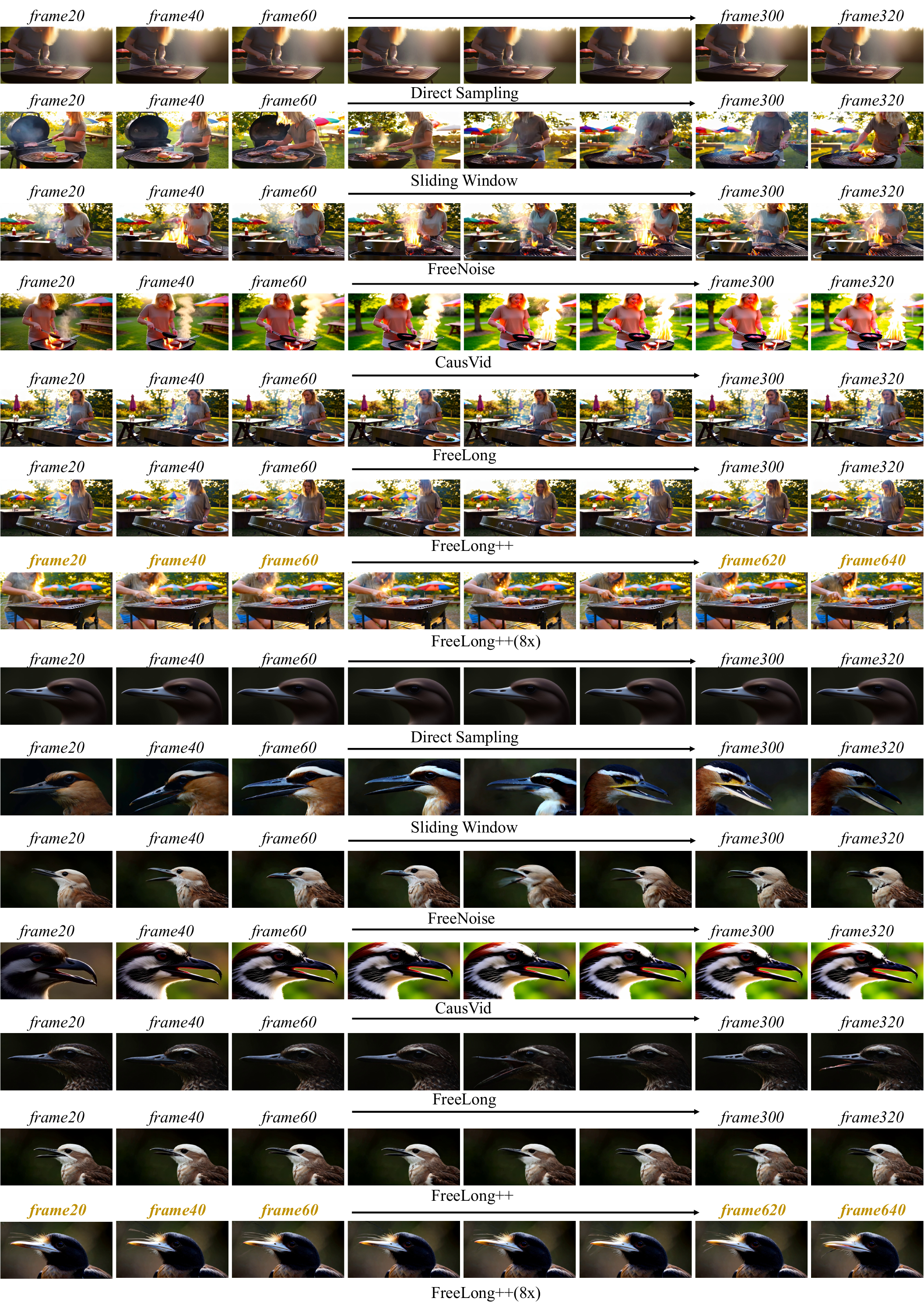}
   \vspace{-0.5cm}
     \caption{
\textbf{Qualitative comparison across models.} All methods generate videos that are 4× the original length, based on the Wan2.1~\cite{wan} model.
         }
   \label{fig:comparison_2}
\end{figure*}

We compare our method against other training-free and training-based approaches for long video generation with generation models, including: (1) Direct sampling, which generates long video sequences directly from short video models; (2) Sliding window, which uses temporal sliding windows~\cite{genlvideo} to process a fixed number of frames at a time; (3) FreeNoise~\cite{freenoise}, which introduces repeated input noise to enhance temporal coherence over long sequences; and (4) CausVid~\cite{causvid}, an autoregressive video generation model fine-tuned from the Wan model.

Tables~\ref{tab:compare_wan} and~\ref{tab:compare_ltx} present quantitative results on Wan~\cite{wan} and LTX-Video~\cite{ltxvideo} models. Advanced DiT video generation models maintain strong motion smoothness and consistency due to variable training video lengths, yet they exhibit lower fidelity in terms of image quality and aesthetics.  Direct sampling leads to high-frequency distortions and significant quality degradation when generating long videos.

Both the sliding-window method and FreeNoise~\cite{freenoise} improve video quality by using fixed temporal attention windows, but still struggle with consistency over long sequences. 
Furthermore, CausVid~\cite{causvid} significantly improves performance on both consistency and fidelity by fine-tuning base model, which require extensive training dataset and computations.

Our FreeLong method outperforms all others, achieving the best scores across all metrics by generating consistent, high-fidelity long videos. Additionally, FreeLong++ further improves image quality and aesthetics by employing multi-band spectral fusion for refined motion dynamics.

\subsection{Ablation Studies}
\begin{figure*}
\centering
   \setlength{\abovecaptionskip}{0.5cm}
   \includegraphics[scale = 0.49]{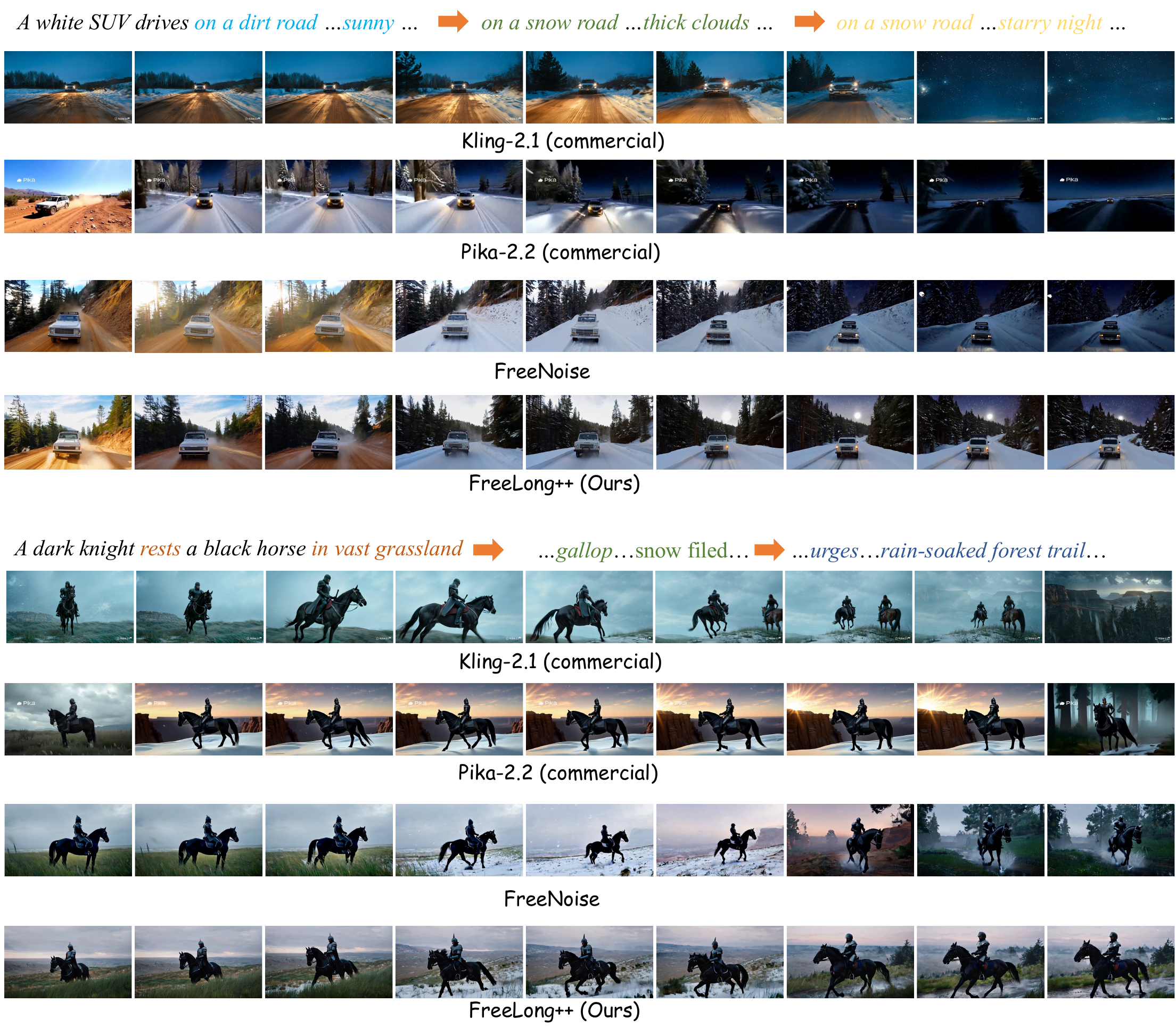}
   \vspace{-.2in}
     \caption{
\textbf{Results of Multi-Prompt Video Generation.} Our method ensures coherent visual continuity and motion consistency across different video segments.
         }
   \label{fig:multiprompt}
\end{figure*}

To evaluate the effectiveness of each component within our FreeLong framework, we conducted a detailed ablation study as summarized in Table~\ref{tab:ablation}. The global-branch approach achieves excellent subjective and background consistency and motion smoothness, but significantly lacks aesthetic and imaging quality. In contrast, the local-branch approach provides improved aesthetic and imaging quality, yet at the cost of lower consistency scores due to limited temporal scope.

Direct addition of global and local branch outputs leads to intermediate consistency but does not effectively improve aesthetic or imaging quality, highlighting the high frequency components degradation caused by naive integration. Our proposed FreeLong method addresses this issue by selectively combining low-frequency global features with high-frequency local features, substantially improving aesthetic and imaging qualities while maintaining high consistency.

The integration of our SpecMix initialization significantly boosts FreeLong's subjective and background consistency respectively, achieving the highest balance across all metrics. Furthermore, the enhanced FreeLong++ further elevates aesthetic and imaging qualities while maintaining superior consistency. Finally, using sparse attention for global-branch notably reduces inference time from 96 seconds to 74 seconds with minimal impact on quality metrics, demonstrating efficient computational performance.

\subsection{Qualitative Comparison}
The synthesis results for each method are presented in Figure~\ref{fig:comparison_2}. In the first row, directly sampling 324 frames from a model trained on 81 frames produces poor results due to high-frequency distortions, resulting in blurred faces and unclear backgrounds. As shown in the second row of Figure~\ref{fig:comparison_2}, using temporal sliding windows generates more vivid videos, but fails to maintain long-range visual consistency, leading to noticeable differences in the subject and background across frames. FreeNoise~\cite{freenoise} aims to improve global consistency by repeating and shuffling initial noise, but still struggles with long-range consistency and suffers from content mutations. CausVid~\cite{causvid} uses auto-regressive architectures to generate coherent video sequences, but is affected by drifting, where visual quality degrades due to accumulated errors over time. In contrast, our method, FreeLong, enforces global constraints during denoising, ensuring temporal consistency and high fidelity across frames. As illustrated in Figure~\ref{fig:comparison_2}, FreeLong produces temporally consistent long videos, outperforming all other methods. Furthermore, FreeLong++ achieves even higher fidelity by using multi-band frequency fusion, better capturing motion dynamics.

\subsection{Multi-Prompt Video Generation}

\begin{figure*}
\centering
   \setlength{\abovecaptionskip}{0.5cm}
   \includegraphics[scale = 0.47]{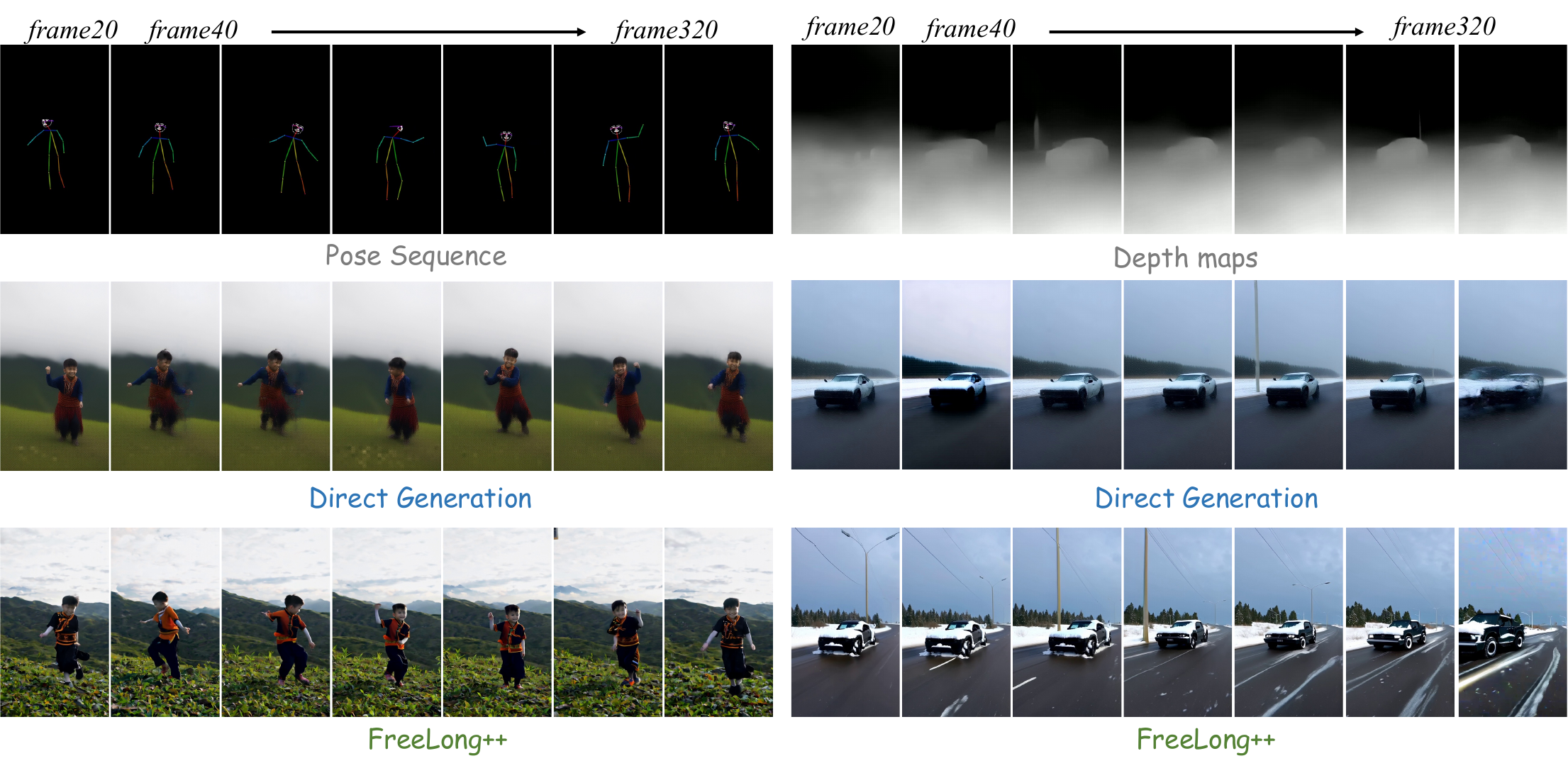}
   \vspace{-.2in}
     \caption{
\textbf{Long Control Sequence.} Long-range video generation under pose (left) and depth (right) guidance. FreeLong++ produces more temporally consistent and semantically faithful outputs than direct generation.
         }
   \label{fig:control}
\end{figure*}

Our method easily extends to multi-prompt video generation by assigning distinct prompts to each video segment. As shown in Figure~\ref{fig:multiprompt}, it maintains coherent visual continuity and consistent motion throughout. For example, a white car drives seamlessly from a dirt road to a snowy road and then into a starry night, all within a unified scene and with smooth transitions. Compared to other approaches, including commercial models like Kling~\cite{kling} and Pika~\cite{pika}, our method achieves superior consistency in scene transitions.
This capability is particularly beneficial for storytelling applications, where maintaining coherence across diverse scenarios is critical. Compare to FreeNoise~\cite{freenoise} that use repeat noise to constrain consistency, our multi-band spectral fusion framework adapts to diverse scenes and temporal complexities, producing videos that are both visually harmonious and temporally logical. In contrast to other systems, our method avoids abrupt or disjointed transitions.

\subsection{Long-Range Control Capability}
FreeLong++ excels at long-range video control by conditioning generation on structured signals such as pose sequences or depth maps over hundreds of frames. As shown in Figure~\ref{fig:control}, our method faithfully adheres to long-duration control signals, preserving consistent motion semantics and scene layout throughout the video. In contrast, direct generation often leads to content drift, identity collapse, or spatial distortion over time. FreeLong++ effectively maintains subject fidelity and background stability across extended sequences, demonstrating its robustness to long-range control signals. This ability is critical for applications like motion-guided synthesis or camera-path conditioning, where fine-grained control must be preserved across the entire video.


\section{Conclusion}
We propose FreeLong++, a training-free framework designed to effectively overcome frequency distortion challenges encountered when extending short-video generative models to longer sequences. By identify high-frequency degradation as a critical limitation, we introduce a multi-band spectral attention mechanism that adaptively integrates temporal features across multiple frequency bands. Specifically, FreeLong++ first employs a multi-window attention module to separately capture video dependencies at distinct temporal scales. Subsequently, it conducts multi-band spectral fusion, systematically fuse these temporal features from low to high frequencies in the spectral domain. This approach significantly enhances temporal consistency and visual fidelity, all without requiring additional training. Our method can be seamlessly integrated into existing diffusion-based video generation models and demonstrates robust performance, consistently producing high-quality long videos across various tasks and model architectures.




%


\ifCLASSOPTIONcaptionsoff
  \newpage
\fi



\small{
\bibliographystyle{IEEEtran}
\bibliography{reference}
}
\vspace{-10mm}

\newpage

\end{document}